\documentclass[conference]{IEEEtran}
\IEEEoverridecommandlockouts
\usepackage{cite}
\usepackage{amsmath,amssymb,amsfonts}
\usepackage{algorithmic}
\usepackage{graphicx}
\usepackage{textcomp}
\usepackage{xcolor}
\usepackage{makecell}
\usepackage{graphics}
\usepackage{float}
\usepackage{subfigure}
\usepackage{cite}

\def\BibTeX{{\rm B\kern-.05em{\sc i\kern-.025em b}\kern-.08em
    T\kern-.1667em\lower.7ex\hbox{E}\kern-.125emX}}
\begin{document}

\title{3DFusion, A real-time 3D object reconstruction pipeline based on streamed instance segmented data
}

\author{\IEEEauthorblockN{1\textsuperscript{st} Xi Sun}
\IEEEauthorblockA{\textit{University of Victoria}\\
Victoria, Canada \\
nathansun@uvic.ca}
\and
\IEEEauthorblockN{2\textsuperscript{rd} Derek Jacoby}
\IEEEauthorblockA{\textit{University of Victoria}\\
Victoria, Canada \\
derekja@uvic.ca}
\and
\IEEEauthorblockN{3\textsuperscript{th} Yvonne Coady}
\IEEEauthorblockA{\textit{University of Victoria}\\
Victoria, Canada \\
ycoady@uvic.ca}
}

\IEEEoverridecommandlockouts

\maketitle
\IEEEpubidadjcol

\begin{abstract}
This paper presents a real-time segmentation and reconstruction system that utilizes RGB-D images to generate accurate and detailed individual 3D models of objects within a captured scene. Leveraging state-of-the-art instance segmentation techniques, the system performs pixel-level segmentation on RGB-D data, effectively separating foreground objects from the background. The segmented objects are then reconstructed into distinct 3D models in a high-performance computation platform. The real-time 3D modelling can be applied across various domains, including augmented/virtual reality, interior design, urban planning, road assistance, security systems, and more. To achieve real-time performance, the paper proposes a method that effectively samples consecutive frames to reduce network load while ensuring reconstruction quality. Additionally, a multi-process SLAM pipeline is adopted for parallel 3D reconstruction, enabling efficient cutting of the clustering objects into individuals. This system employs the industry-leading framework YOLO for instance segmentation. To improve YOLO's performance and accuracy, modifications were made to resolve duplicated or false detection of similar objects, ensuring the reconstructed models align with the targets. Overall, this work establishes a robust real-time system with a significant enhancement for object segmentation and reconstruction in the indoor environment. It can potentially be extended to the outdoor scenario, opening up numerous opportunities for real-world applications. 

\end{abstract}


\section{Introduction}
One of the main research directions of computer vision lies in how to give machines the ability to analyze images intelligently. The semantic understanding of scenes is crucial for achieving interactivity in many areas like urban planning, interior design, AR/MR, telemedicine, etc. However, semantic analysis of 2D images can not fulfil the requirements for these applications. We also need to understand the spatial layout of objects in a 3D environment. The detection and reconstruction of 3D objects have become an important aspect of computer vision research.

With the development and maturity of 3D imaging technologies, such as structured light measurement, laser scanning, and ToF, the 3D coordinates of every visible object can be accurately and quickly acquired to construct 3D data of the scene. 3D data that contains the RGB-depth and position information can help outline the shape of the object to provide a better representation of the surrounding environment. In some application scenarios that require responsiveness, such as robotics, augmented reality, autonomous driving, remote sensing, etc., real-time detection and reconstruction of 3D scenes and objects have a broad prospect.

3D data are often obtained directly from sensors in different forms such as depth maps, point clouds, and grids. Among them, point cloud data is convenient to obtain and has the capability to store multi-dimension information. Point cloud becomes the mainstream of 3D segmentation research in recent years. The rapid development of deep learning has brought great advantages to point cloud segmentation. We can use machine learning algorithms (e.g., deep neural networks or clustering methods) to perform various image segmentation tasks, such as semantic segmentation, object detection, and instance segmentation.

However, unlike the organized pixels in 2D images, point cloud data is disordered, which makes it difficult to apply convolution directly to obtain local correlations between 3D points. Furthermore, the point cloud data are often non-uniformly distributed. The density of the point cloud in different areas is often unequal, which makes it difficult to sample the data points for feature extraction. The deformation of objects in 3D space is more complex than that of 2D images due to the non-rigid body deformation and the affine transformation in three dimensions.

In many practical computer vision scenarios, such as robotics or AR/VR, the scene is always dynamic and ever-changing which requires real-time segmentation. However, 3D segmentation requires processing every point or voxel, which causes a slow processing speed. The complexity of the 3D segmentation algorithms also makes the application computationally expensive in real-time scenarios. 

To avoid these weaknesses, we propose a real-time segmentation and reconstruction pipeline based on 2D segmentation model. In contrast to the direct segmentation of 3D point clouds, our pipeline performs segmentation on real-time 2D serialized images from a multi-view scanning, and then annotates and reconstructs the objects of interest in 3D representation. The annotation of objects improves the user's understanding of the semantics of the scene and enables 3D interaction with objects.

\section{Related Work}
\subsection{Data Preparation}
Spatial data is acquired by sensors which produce various representations including RGB images, depth or grayscale images, and Lidar images. But it is hard to intuitively derive the 3D shape from 2D images. In order to recognize 3D objects, the relevant data needs to be segmented from the background and reconstructed into spatial models as cognitive references. There are usually two methods: reconstruction based on RGB images acquired from multiple views of the object\cite{schoenberger2016sfm}]\cite{wang2021deep} \cite{nerfstudio} and reconstruction based on depth images \cite{izadi2011kinectfusion}\cite{kerl2013robust}\cite{whelan2015elasticfusion}\cite{choi2015robust} which have had significant development in recent years. 

The color information of the object can change greatly in a wide range of variations due to different illumination and reflection. Moreover, in RGB images, there are substantial obstacles in the detection of overlapping objects. It is hard to identify the objects completely from the background. So, the uncertainty of the color and occlusion in RGB images challenges the reconstruction. 

In contrast, a depth image is formed by the distance from each pixel to the camera and can be obtained by various methods such as structured light scanning, stereo reconstruction or ToF (time-of-flight) capture. The 'depth' is independent of texture and perspective. Depth images can eliminate the random changes in color and texture caused by lighting and shadow. 

The grayscale aligns with the field-of-view direction. By using depth information, 3D models can be generated by capturing the geometry of the objects in the scene regardless of the object occlusion or partial overlap. Even if there is occlusion, the objects can still be distinguished by the depth difference between the front and back objects.

The quality of 3D reconstruction based on depth images depends on the accuracy of the depth information. However, there are many challenges in obtaining high-quality depth images, such as noise, occlusion, and limited sensing range. As consumer-grade RGB-D cameras continue to evolve, such as Microsoft Kinect, Intel RealSense, PrimeSense, etc., obtaining depth information has become less challenging over time. The algorithms for large-scale 3D scene reconstruction of images captured by low-cost RGB-D cameras have been extensively developed \cite{izadi2011kinectfusion}\cite{whelan2015elasticfusion}\cite{choi2015robust}\cite{runz2018maskfusion}\cite{ORBSLAM3_TRO}. These methods have been extended to allow remote exploration of 3D scenes during on-site reconstruction using VR devices \cite{holler2021automatic}.

\subsection{Image segmentation and object detection}
With the development of convolution network architectures, efficient and effective object detection and semantic annotation have become possible. There were great achievements in object detection and segmentation of 2D images \cite{redmon2016you}\cite{jain2023oneformer}\cite{cheng2021perpixel}\cite{carion2020endtoend}\cite{shafiee2017fast}\cite{ren2015faster}\cite{liu2016ssd}\cite{redmon2016you}\cite{he2017mask}. Most network structures, such as Fast R-CNN \cite{girshick2015fast}, Faster R-CNN \cite{ren2015faster} use two-component systems to perform classification and localization separately. Candidate regions are extracted and subjected to deep learning to perform classification. Fully convolution networks such as Mask-RCNN \cite{he2017mask}, use deconvolution to specify an instance mask for each pixel, extending single object detection to semantic and instance segmentation. Deep-learning-based regression methods, such as YOLO \cite{redmon2016you} \cite{shafiee2017fast} and SSD (Single Shot Detectors) \cite{liu2016ssd}\cite{lin2017focal}, solve the classification and localization problems at once. The latter two methods are commonly used in modern object detection and segmentation research and products. With the reduced difficulty of RGB-D data acquisition, object detection and segmentation of 2D images are extended to the 3D domain \cite{song2014sliding}\cite{song2017semantic}. These studies aim to infer the bounding box of 3D objects from RGB-D frames based on deep learning of 2D images.

Regarding 3D object detection and segmentation, the multi-view method leverages RGB-D video streams to classify objects based on a set of segmented 2D images. These images are projected onto 3D space to enable 3D object segmentation \cite{su2015multi}\cite{kalogerakis20173d}. 
Thanks to the latest developments in 3D deep learning, it's now possible to establish convolution operators on networks using voxels and points. With the advent of 3D shape datasets \cite{chang2015shapenet}\cite{Matterport3D}\cite{Armeni_2016_CVPR}\cite{Behley_2019_ICCV}\cite{hackel2017semantic3dnet}\cite{tan2020toronto3d} and point-based networks, 3D convolution networks have been used for 3D object classification \cite{qi1706deep}\cite{hu2020randla}\cite{zhou2018voxelnet}\cite{shi2019pointrcnn}, semantic segmentation \cite{qi1706deep}\cite{li2018pointcnn}\cite{hu2020randla}\cite{thomas2019kpconv}\cite{kirillov2020pointrend}, and instance segmentation \cite{zhang2020instance}\cite{mo2019partnet}\cite{zhao2020jsnet}. However, most point-based methods rely on expensive adjacency search mechanisms, such as KNN \cite{1053964}. While the use of point-voxel-based joint representation can enhance segmentation efficiency, it is only effective for specific datasets. For example, RANDLA-Net achieves 77.4\% A-IOU on Semantic3D dataset, but only 55\% IOU on SematicaKitti \cite{hu2020randla}. 

The above 3D segmentation methods are usually used to deal with static, large-scale point cloud scenes. Real-time segmentation of point clouds in dynamic, unknown scenes has been little studied. Therefore, when it comes to object segmentation of real-time 3D scenarios, 2D classification and segmentation algorithms are currently far superior to 3D-based methods in terms of efficiency and accuracy.

\subsection{3D Reconstruction}

3D reconstruction is the process of constructing a 3-dimensional model of an object or a scene from 2D images or point cloud data. This is achieved by estimating the depth and location and capturing the geometry and appearance of objects in the scene. Early 3D reconstruction techniques typically used 2D RGB images as input. However, the reconstructed 3D models were usually incomplete and less realistic due to color variations and occlusion. With the emergence of consumer-grade depth cameras, 3D scanning and reconstruction technology have been developed rapidly. 

\begin{figure}
    \centering
    \includegraphics[width=.6\linewidth]{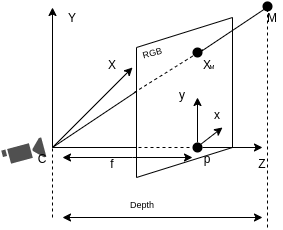}
    \caption{Coorespondence between RGB and Depth Images}
    \label{fig:RGBD}
\end{figure}

RGB-D data scanned by the depth camera consists of the RGB and depth information. Each pixel in an RGB image corresponds to a point in the depth image. The depth value represents the true distance from each point to the vertical plane where the depth camera is located. Figure \ref{fig:RGBD} illustrates the relationship between an RGB image and a depth image. With the camera position as the origin point, the direction the camera is facing as the Z-axis, and the two axes of the camera's vertical plane as the X and Y-axes, we can establish the local 3D coordinate system of the camera. For point M, the depth camera is able to obtain its imaging point $X_M$ in the RGB image and the distance from M to the vertical plane (i.e., XY plane). This distance is the depth value of M. Given the focal length of the camera, we can easily get the 3D coordinates of M in the local coordinate system via a simple trigonometric formula. Each pixel in the RGB image is able to convert to a 3D point, thus, we can derive a point cloud model from a RGB-D image. 

\addtolength{\topmargin}{0.05in}

There is another critical matter in the reconstruction process - the estimation of camera positions. The camera position changes consistently. Each camera position corresponds to a different frame and local coordinates. However, to reconstruct the model, it must be located in a unique coordinate system - world/global coordinates. Thus, we need to find the relationship between the local coordinate system of the camera and the world coordinate system in each frame. In the field of computer vision and intelligent robotics, this problem is the classical "Simultaneous localization and mapping" (SLAM) localization problem.

KinectFusion, a classical method for 3D reconstruction proposed by Izadi et al. in 2011 \cite{izadi2011kinectfusion}, uses the iterative closest point (ICP) method to solve the above problem. ICP calculates the transfer matrix from the original data to the target data based on the correspondence between their data points and minimizes the sum of the distances from all target data points to the tangent plane where their corresponding original data points are located.

However, ICP algorithm in KinectFusion only uses the data in 3D space and does not take into account the RGB data information. In addition, ICP must be based on the premise that the frame rate is high and the difference between two adjacent frames is minimal. This method has limitations to estimate the camera position, particularly in scenes with large planes (e.g., walls, ceilings, and floors) where it can introduce significant errors. However, considering real-time and stability, this ICP-based framework is a commonly used system for estimating camera positions.

Another technique for constructing 3D models from 2D images is called SFM (Structure from Motion) \cite{schoenberger2016mvs}\cite{schoenberger2016sfm}. The process involves capturing multiple images of an object or scene from different angles and calculating their corresponding camera positions. This information can then be used to generate 3D models, point clouds, and other representations of the 3D scene. SFM is commonly used in applications such as 3D reconstruction of cultural sites, content creation for virtual reality and augmented reality, etc.

We generally consider the basic theory of SFM and SLAM to be similar, both being based on multi-view geometry. But there are still some differences. When using SFM reconstruction, the image inputs do not need to be in a specific order. The reconstruction may take more time to process but provides more accurate results. SLAM, on the other hand, is developed from ICP algorithm, which requires strictly ordered images and minimum transformation between consecutive frames, so they can be processed online in real-time. SLAM is more often used in small computers and embedded devices, such as the Nvidia Jetson-Nano or Rasberry-Pi, for mobile purposes.

\section{Methodology}
Although a number of papers have integrated SLAM and CNN to implement object detection and 3D reconstruction \cite{whelan2015elasticfusion} \cite{runz2018maskfusion} \cite{nakajima2018efficient} \cite{mccormac2017semanticfusion}, most of these cannot cope with the real-time camera input. Their reconstruction process is isolated from the segmentation. They lack the ability to quickly and accurately detect 3D objects in a short time. As a result, this hinders the ability to remotely explore and interact with 3D objects.

\begin{figure}
    \centering
    \includegraphics[width=.8\linewidth]{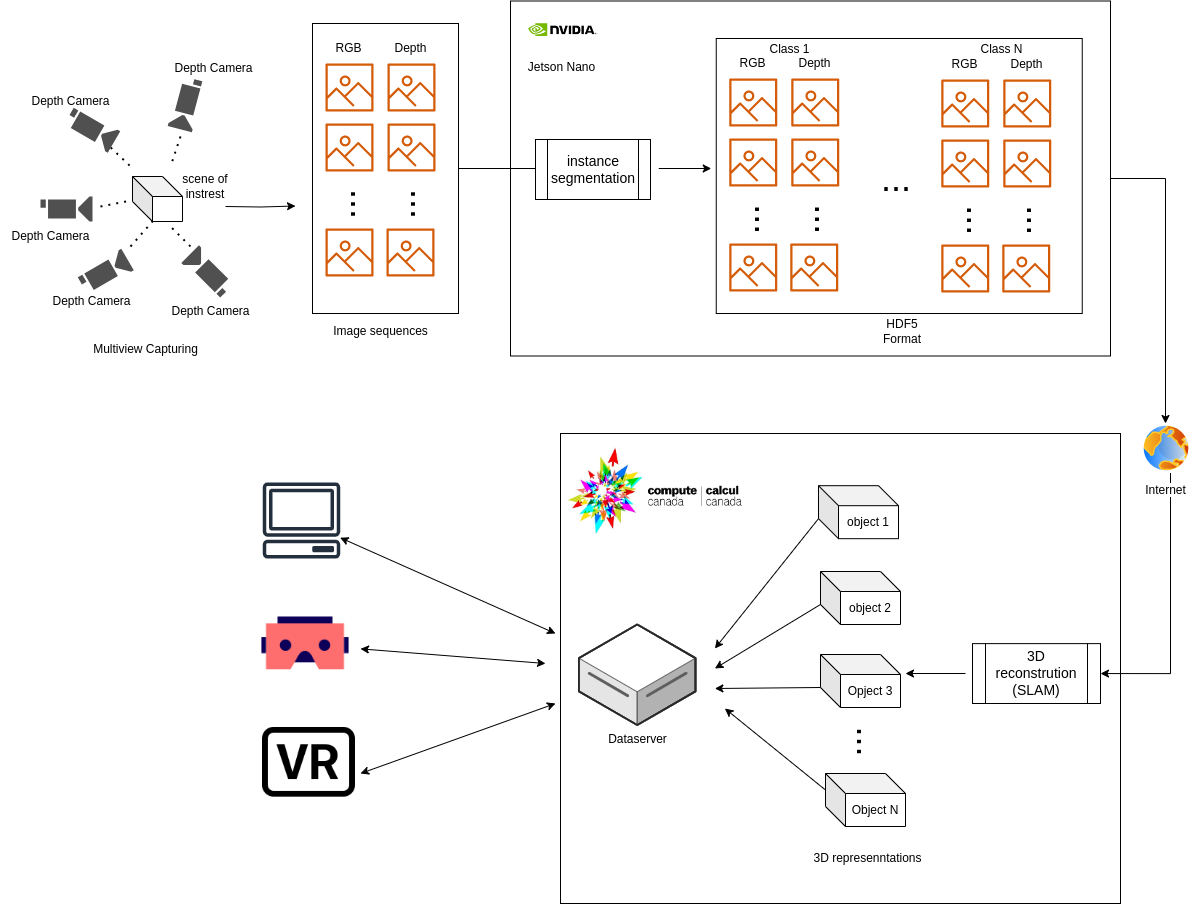}
    \caption{Pipeline of Segmentation and Reconstruction}
    \label{fig:pipeline}
\end{figure}

To overcome these limitations, in the following sections, we propose a pipeline (\ref{fig:pipeline}) for real-time 3D object reconstruction. Our approach integrates and enhances the methodologies from different research areas, in particular, 2D instance segmentation and 3D reconstruction on segmented RGB-D images using SLAM. The whole pipeline consists of the following steps.

First, we utilize Intel's Realsense RGB-D camera installed on a Jetson Nano kit to record real-time images and run a preliminary instance segmentation using YOLOV8 \cite{Jocher_YOLO_by_Ultralytics_2023}. Instance segmentation not only identifies the semantic objects within the image but also inference a mask for each object. So we can filter all identified objects out of the background with unique labels. 

In the second step, some objects of interest are segmented from RGB-D images using instance masks and categorized into multiple image groups by the labels. These images and their metadata which contain the intrinsic camera parameters and classification information are formatted as HDF5 data and transferred to our server located in Compute Canada \cite{ccdb} for reconstruction and storage. Compute Canada is a high-performance computing platform provided by the Digital Research Alliance of Canada. It provides powerful infrastructure of computation, such as high-performance GPU, and Terabyte grade memory. 

There, we perform the third step "3D reconstruction" to generate classified representations of objects using SLAM. These representations are usually exported and stored as major 3D files, such as ply, obj, gltf, 3DTiles, etc, accordingly to different visualization and interaction purposes.

\addtolength{\topmargin}{0.05in}
\subsection{Data Collection}
We use Intel Realsense D455 as the data acquisition device. As a portable depth camera, Realsense can be attached to any small computer system, like Nvidia Jetson Nano, Rasberry Pi, FPGAs, and installed on mobile terminals, such as robots and vehicles to collect color and depth images simultaneously. Its real-time performance to collect images and adaptability to illumination is suitable for indoor and outdoor uses. Its basic parameters are listed in table \ref{tab1}. It can scan a range of up to 6m, which is the largest in the Realsense series. The frame rates for both depth and RGB are sufficiently high to facilitate SLAM processing.

\begin{table}
    \caption{Realsense Specification}
    \begin{center}
    \resizebox{.8\columnwidth}{!}{%
    \begin{tabular}{|c|c|c|c|}
    \hline
    \multicolumn{2}{|c|}{\textbf{\makecell{Depth \\ Parameters}}}&\multicolumn{2}{|c|}{\textbf{\makecell
    {RGB \\Parameters}}} \\
    \hline
    \textbf{\textit{\makecell{Depth FOV \\ (H x V)}}} & {87° × 58°} & \textbf{\textit{\makecell{RGB Resolution \\ and Frame Rate}}}& {\makecell{1280 × 800 \\ at 30 fps}} \\
    \hline
    \textbf{\textit{\makecell{Depth Resolution}}} & \makecell{Up to \\ 1280 × 720} & \textbf{\textit{\makecell{RGB Sensor \\ FOV}}}& {90° × 65°} \\
    \hline
    \textbf{\textit{\makecell{Depth Accuracy}}} & \makecell{ $<$ 2 $\%$ \\at 4 m} & {} & {}\\ 
    \hline
    \textbf{\textit{\makecell{Depth Frame \\ Rate}}} & \makecell{Up to \\ 90 fps} & {} & {} \\
    \hline
    \textbf{\textit{Ideal Range}} & \makecell{.6m to \\ 6 m} & {} & {} \\
    \hline
    \end{tabular}%
    }
    \label{tab1}
    \end{center}
\end{table}

Our data collection platform is a customized Jetson Nano from Seeed Studio$\copyright$. This device has an ARM architecture and 4GB memory, making it capable of running small-size segmentation models. We implemented a Realsense client to stably scan the scene at an average 30 FPS which guarantees the accuracy of the point cloud registration in the later SLAM processing. 

However, the camera generates massive and redundant images at high FPS. When the camera moves continuously and steadily, appropriately reducing FPS can optimize the network load without impacting the result of reconstruction. Normally, Using keyframe to represent its nearby frames is a common method to reduce the number of frames in some SLAM algorithms, like ORB-SLAM \cite{ORBSLAM3_TRO}, which reduces the computation of these complex algorithms. 

In our clients, we minimize the number of frames for transmission. Typically, we extract the keyframe from each stationary frame, such as every 10 frames. This method is not efficient. When the motion is slow, too many similar keyframes are selected, which is unnecessary. On the other hand, when the motion is fast, important frames are likely to be missed. So we consider the spatial change from the adjacent keyframe. We calculate the relative movement (Euclidean distance \eqref{eq1}) between the two frames according to their poses. The new keyframe will be created when a movement reaches a threshold. We calculate the similarity \eqref{eq2} of the frame to the previous keyframe. We discard it if it has a high similarity, but mark it as keyframe otherwise. In \eqref{eq1}, \eqref{eq2}, matrices of frames are converted to two-dimensional vectors to facilitate calculation. The extraction of frames significantly accelerates the transmission of valid frames over the network.

\begin{equation}
\begin{aligned} 
A=(a_1,a_2,...a_n), B=(b_1,b_2,...b_n) \\
d(A,B)=\sqrt{\sum_{i=1}^{n}\sum_{i=1}^{n}(a_ij - b_ij)^2}
\label{eq1}
\end{aligned}
\end{equation}

\begin{equation}
\begin{aligned} 
A=(a_1,a_2,...a_n), B=(b_1,b_2,...b_n) \\
cos(A,B)=\frac{\sum_{1}^{n}(a_i \times b_i)}{\sqrt{\sum_{1}^{n}a_i^2} \times \sqrt{\sum_{1}^{n}b_i^2}}
\label{eq2}
\end{aligned}
\end{equation}

\subsection{Realtime 2D Instance Segmentation}
After capturing and sampling the RGB-D images, we apply the 2D instance segmentation to segment the objects of interest in the frontend device - Jetson Nano. Since it only has 4GB memory, many large models can not be loaded. YOLO (You Only Look Once) \cite{redmon2016you} is a compromise choice for our segmentation task. It is an advanced object detection and image segmentation model that offers multiple pre-trained models for different tasks. 

As of the time this paper was written, YOLOV8 is the latest version of the YOLO series. The largest segmentation model YOLOv8x-seg is only 137MB with the highest mAP, but the inference speed is correspondingly slow. To compromise the accuracy and speed to achieve real-time segmentation on continuous frames, we choose YOLOv8s-seg model instead of YOLOv8x-seg for our purpose on Jetson Nano. According to a disclosure of the YOLOv8 models' performance \cite{Jocher_YOLO_by_Ultralytics_2023}, the YOLOv8s-seg is around 22MB but can perform the inference at 155ms/img on CPU and 1.47ms/img on GPU, which is 3 times faster than YOLOv8x-seg. YOLOv8s-seg is trained over the COCO segmentation dataset with images resolution of 640 $\times$ 480. We need to preset the resolutions of both RGB and depth images in Realsense Camera in order to collect the compatible data.

\begin{figure}
    \centering
    \subfigure[]{
    \label{Fig. sub.0}
    \includegraphics[width=3cm,height = 2cm]{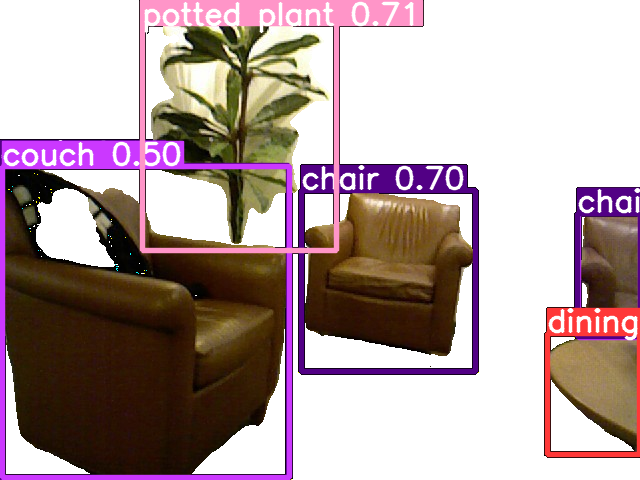}
    }\subfigure[]{
    \label{Fig. sub.1}
    \includegraphics[width=3cm,height = 2cm]{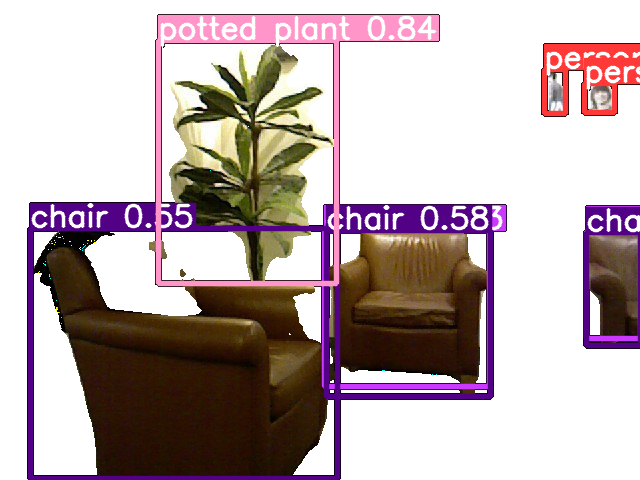}
    }
    \subfigure[]{
    \label{Fig. sub.2}
    \includegraphics[width=3cm,height = 2cm]{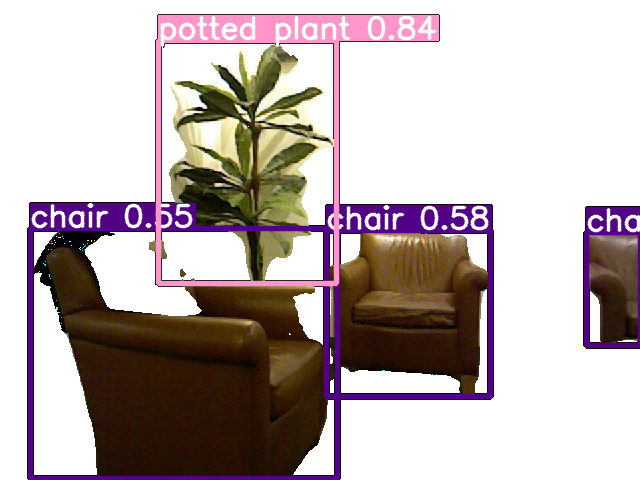}
    }\subfigure[]{
    \label{Fig. sub.3}
    \includegraphics[width=3cm,height = 2cm]{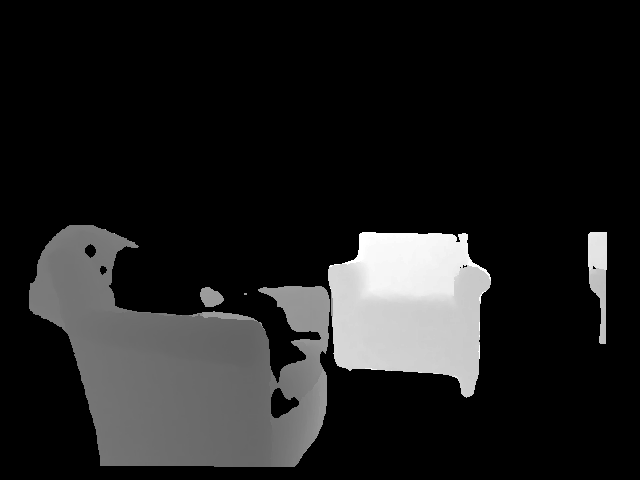}
    }
    \caption{Segmentation Results using YOLO on a Standford Lounge RGB-D Datasets \cite{armeni_cvpr16}}
    \label{fig:YOLOSEG}
\end{figure}

After running the YOLO segmentation, we can get numerous predictions of objects including bounding boxes, object masks and corresponding labels with confidences (Figure: \ref{fig:YOLOSEG}). A mask is a binary matrix that marks the pixels of the identified object in an image. We can apply the masks on original RGB-D images to filter the visible objects out of the background and only preserve the RBG and Depth information of these objects. It is the premise for a precise reconstruction. 

The YOLO is not good at distinguishing similar classifications, such as chair and couch, which contain many common features. To solve this problem, we utilize IOU (intersection of union) coming from NMS (non-maximum suppression) algorithm. NMS is usually used to reduce and merge redundant and overlapping bounding boxes in image segmentation. Borrowing the concept, we calculate the IOU of each mask of the objects \eqref{eq3}. Similar to calculating the area of bounding boxes in NMS, we use the numbers of pixels in the intersection and union of two masks to compute IOU. When the IOU value exceeds a threshold (e.g. 0.45), we discard the mask with fewer pixels. On the contrary, when the IOU is less than the threshold, two actual objects are detected with some overlaps and the intersection of the two objects is assigned to the object with a higher confidence score. In order to reduce the number of candidates, masks that have lower scores (e.g. 0.35) are ignored before the IOU calculation. 

Generally, Realsense Camera exports two types of images, RGB and Depth. Applying segmentation to RGB images produces masks that can be used to select matching points in both RGB and Depth images \ref{Fig. sub.3}. The segmented objects in one image are stored in HDF5 (The Hierarchical Data Format version 5) structure with their RGB, depth information and labels. HDF5 supports complex, heterogeneous data. Its directory-like structure allows us to organize data in customized structures with embedded self-describing metadata \cite{hdf5}. 

\begin{figure}
    \centering
    \includegraphics[width=.8\linewidth]{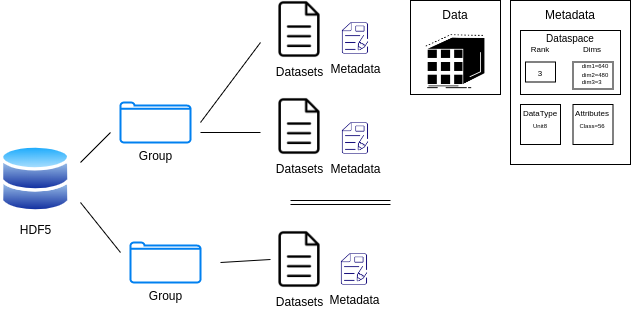}
    \caption{HDF5 structures}
    \label{fig:hdf5}
\end{figure}

In our case, we store RGB and Depth datasets for each object. Each dataset comes with its own metadata which contains the label of the object, data type, and data dimensions. Objects are grouped and stored by categories. The entire structure is shown in Figure \ref{fig:hdf5}. The HDF5 data is then sent to the backend utilizing a Python package "PyDataSocket", which supports the HDF format transmission via TCP socket. There, a further 3D reconstruction will be performed.

\begin{equation}
\begin{aligned} 
Mask_{class}= Masks(IoU_{thres}, C_{class})\\
IoU=\frac{pixels\ of\ overlap}{pixels\ of\ union}
\label{eq3}
\end{aligned}
\end{equation}

\subsection{Real-time Reconstruction Based on SLAM}
\begin{figure}
    \centering
    \includegraphics[width=.8\linewidth]{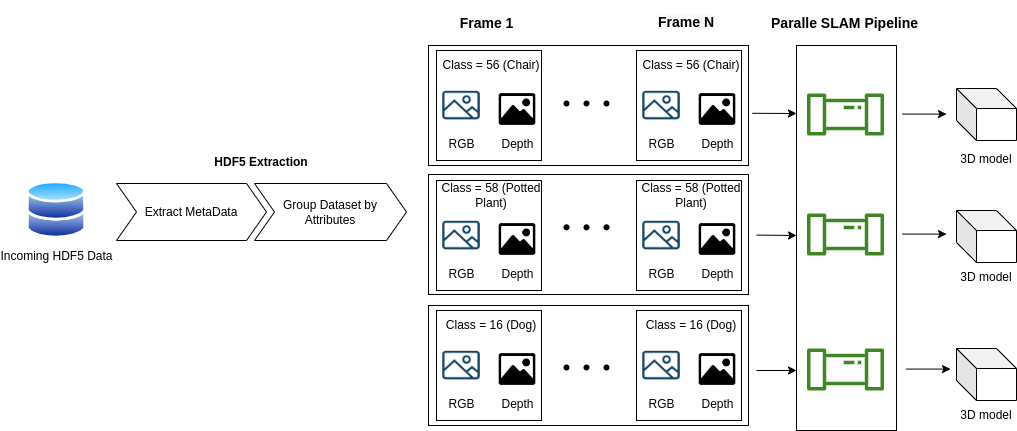}
    \caption{HDF5 Extraction into Classified RGB-D Data}
    \label{fig:hdf5extract}
\end{figure}

In this section, we implement a SLAM-based dense 3D reconstruction pipeline deployed on Compute Canada Cedar Cluster \cite{cedar}. This pipeline continually receives the HDF5 data from the network with a FIFO buffer. With the labels in metadata, the pipeline can group the datasets from the HDF5 data by object categories in order. The reorganized data is used to generate 3D models of interest in the captured scenes (Figure \ref{fig:hdf5extract}). As the incoming data is categorized by various labels, multiple threads can be created in parallel according to the number of categories for reconstruction, taking advantage of the powerful infrastructure of Compute Canada. 

In the reconstruction step, we utilize an open source python library -  Open3d \cite{Zhou2018} to perform a dense RGB-D SLAM and fast volumetric reconstruction using frame-to-model tracking based on point cloud registration \cite{8237287}, \cite{7299195}. It provides numerous functionalities to handle point cloud data and model reconstruction, making it ideal for our intended use.

\begin{figure}
    \centering
    \includegraphics[width=.8\linewidth]{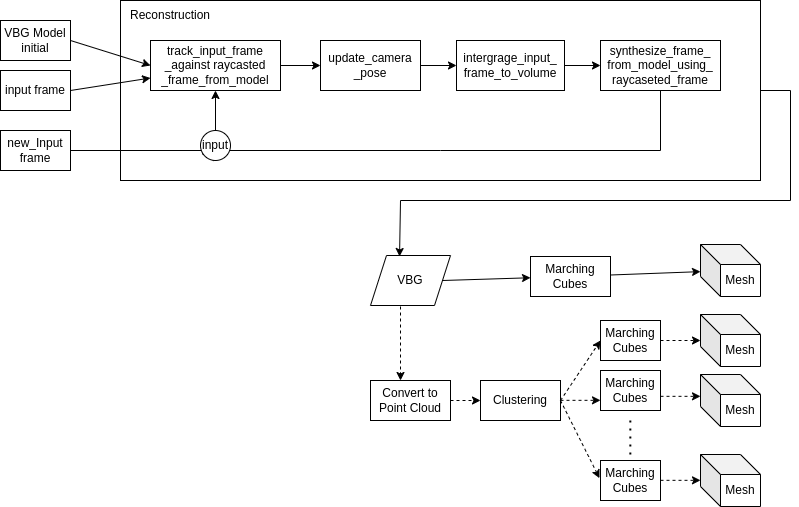}
    \caption{Reconstruction Processing}
    \label{fig:reconstruction}
\end{figure}

An individual reconstruction (Figure \ref{fig:reconstruction}) starts from the initialization of a model and an input frame. The model is created upon an Open3D data structure - Voxel Block Grid (VBG) and maintained all the time in the whole processing with the new incoming input frames. A frame-to-model tracking loop is executed to update the synthesized frame via volumetric ray-casting from the model and perform the RGB-D odometry between the input frame and the synthesized frame. The final model is generated by TSDF (Truncated Signed Distance Function) integration offered by Open3D, which integrates depth inputs into the VBG according to the camera poses. The surfaces of objects can be extracted from the VBG model using the marching-cubes algorithm\cite{lorensen1998marching}. 

\addtolength{\topmargin}{0.05in}

\begin{figure}
    \centering
    \subfigure[]{
    \label{Fig. sub.4}
    \includegraphics[width=3cm,height = 2cm]{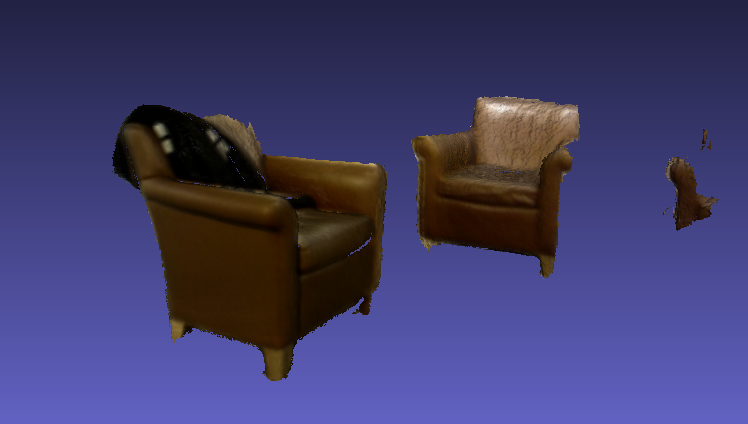}
    }\subfigure[]{
    \label{Fig. sub.5}
    \includegraphics[width=3cm,height = 2cm]{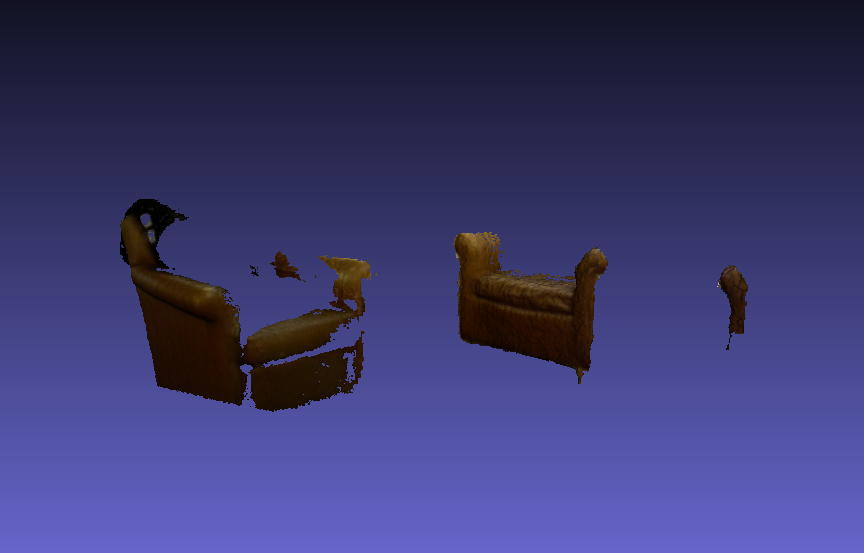}
    }
    \caption{Reconstructed Assembly of Chairs (a): normal extraction, (b): under extraction}
    \label{fig:3dmodels}
\end{figure}

We perform each reconstruction on a particular category of RGB-D image sequences, which results in a collection of objects with the same label, for example, several chairs are reconstructed as a whole \ref{Fig. sub.4}. If we want to split the assembly into individual objects, we need to find some clustering algorithms for VBG models before running marching-cubes reconstruction (shown as dotted lines in Figure \ref{fig:reconstruction}). 

\begin{figure}
    \centering
    \subfigure[]{
    \label{Fig. sub.7}
    \includegraphics[width=.6\linewidth]{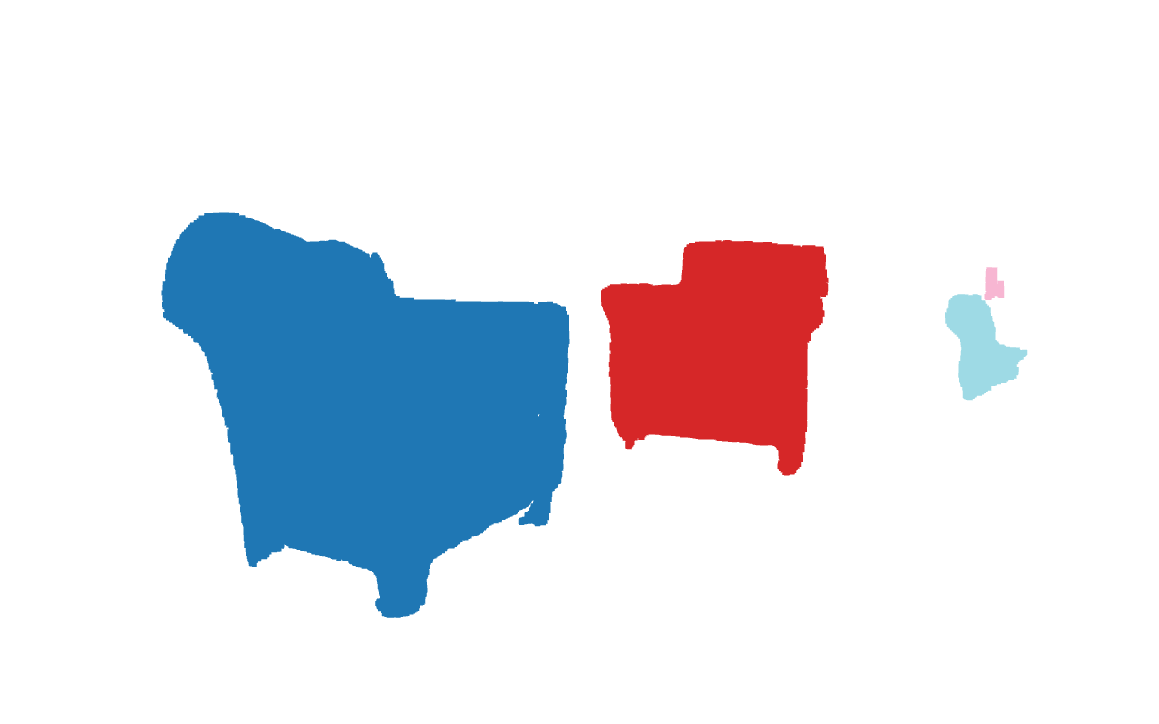}
    }
    \subfigure[]{
    \label{Fig. sub.8}
    \includegraphics[width=3cm,height = 2cm]{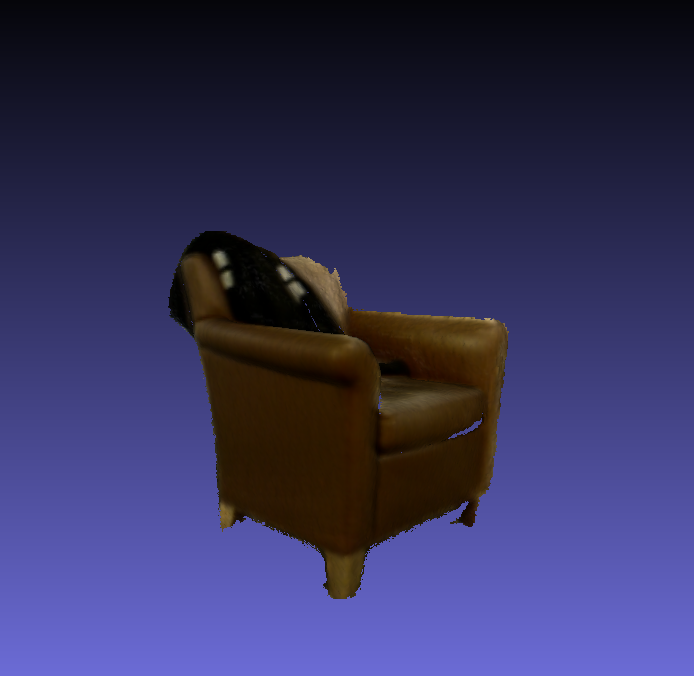}
    }\subfigure[]{
    \label{Fig. sub.9}
    \includegraphics[width=3cm,height = 2cm]{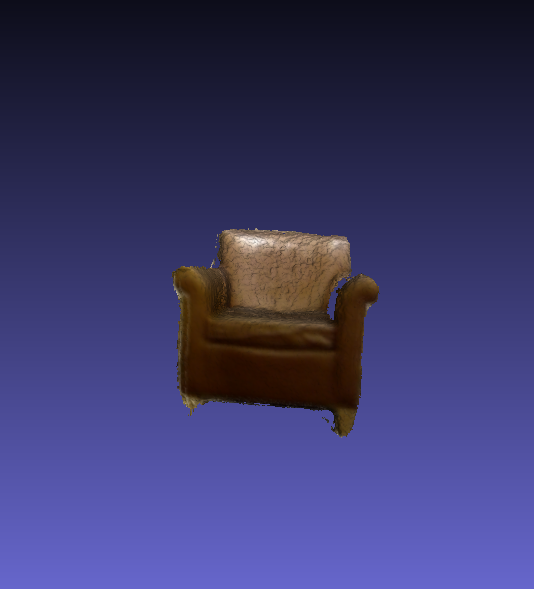}
    }
    \caption{(a)Point Cloud Segmentation using RANSAC and DBSCAN, Visualized by Coloring, (b)(c) Individual Reconstruction by Marching Cubes}
    \label{fig:segmented_pc}
\end{figure}

Clustering is a large concept in 3D segmentation. Many mature algorithms have been developed, including Random Sample Consensus(RANSAC) \cite{fischler1981random}, Density-based Spatial Clustering of Applications with Noise (DBSCAN) \cite{ester1996density}, Euclidean Cluster Extraction, Region Growing Segmentation, Min-Cut Based Segmentation, and more algorithms based on deep learning \cite{guo2020deep}. However, these algorithms all act on point clouds, so we need to convert the VBG model into a point cloud in advance. To balance the accuracy and algorithm complexity, we applied both the RANSAC and DBSCAN on the point clouds and then perform the marching-cubes algorithm on each segmented object for individual surface reconstruction (Figure \ref{Fig. sub.8} and \ref{Fig. sub.9}).

\section{Results and Evaluation}
The end-to-end processing time of a frame consists of the time of RGB-D image capture and segmentation, data transmission over the network, and frame-to-model tracking in reconstruction. The maximum frame rate of the camera is configured as 30 FPS. However, the actual frame rate varies since we allow new frame capture only when the previous one is segmented and transmitted to the server successfully. The segmentation is processed approximately at 600 FPS (1.47ms/img) on GPU. The frame integration and model updating in SLAM on the server are fast. The pipeline can achieve the real-time performance as long as the network is fast and stable.  

we will evaluate the pipeline in the following aspects: efficiency of data preparation and transmission, and performance of the 3D reconstruction. 

\subsection{Data preparation and transmission}

The Realsense RGB-D camera has the capability to scan the surroundings at a maximum speed of 90 FPS, with 640 by 480 resolution. And we set it to 30 FPS. In our implementation, the camera starts to capture the next frame when the server receives the previous one to ensure the image sequence. So the data transmission speed over the network impacts the actual scanning FPS. The size of the output HDF5 data generated from the segmentation of one frame is around 320kb before compression. Theoretically, to reach the maximum capturing frame rate, the network upload capability needs to be higher than 10Mbps. However, the transmission speed is affected by many factors, like network protocols, internet providers, and the quality of the router. 

The reconstruction is performed in Computer Canada, which requires us to transmit the data via an unpredictable internet. In our experiment, the average frame transmission rate is 8 FPS. Thus, we can only capture and process 8 images per second in the reconstruction. In the indoor scene, the camera is usually handheld and the change among frames is minor. Our reconstruction still works properly and reliably with such a low frame rate \ref{Fig. sub.4}. 

Moreover, as we discussed in section 3.A, we perform frame extraction before the 2D segmentation to reduce the number of frames to be processed. However, oversampling of the raw images may bring odometry drift and information loss, which will distort the reconstructed model (Figure\ref{Fig. sub.5}). To avoid distortion, we minimize the sampling or even skip it when the frame rate is low. 

As described before, one object might be classified into multiple categories. For example, in the image (b) in Figure \ref{Fig. sub.2}, 7 objects are detected while there are only six actually. One object is classified as both couch and chair with different confidence scores as listed in table \ref{tab3}. With the strategies discussed in section 3.B, we eliminate identifying one object as multiple ones mistakenly as shown in Figure \ref{Fig. sub.3}.

\begin{table}
    \caption{Detected objects and confidences}
    \begin{center}
    \resizebox{.5\columnwidth}{!}{%
    \begin{tabular}{|c|c|}
    \hline
    \textbf{\textit{Classification}} & \textbf{\textit{Confidence}} \\
    \hline
    couch &0.27 \\
    \hline
    person &0.28 \\
    \hline
    person &0.33 \\
    \hline
    couch &0.33 \\
    \hline
    chair &0.40 \\
    \hline
    chair &0.55 \\
    \hline
    chair &0.58 \\
    \hline
    \end{tabular}
    \label{tab3}%
    }
    \end{center}
\end{table}

 Additionally, HDF5 has the capability to compress the data when using the Python library H5PY. There are 10 compression levels of 'gzip' option. The compression rate is around 15\% at the medium compression level and 30 \% at the highest level but with a longer processing time. The compression can increase transmission FPS which brings an increase in capturing FPS. However, this might result in a more aggressive sampling. To increase the overall pipeline FPS, we can adjust the thresholds of sampling. With the balance between compression and sampling, we observe an increase in the overall pipeline frame rate from an average of 8 FPS to 10 FPS in our lab environment. Therefore, the pipeline can be more responsive and achieve better performance in real-time.

\subsection{3D reconstruction}
The amount of time it takes for reconstruction is short compared to the whole processing time of the pipeline. The system takes a while to initialize the of VBG model when receiving the first frame. After the initialization, the average processing time is less than 20ms per frame (Table \ref{tab4}). The reconstruction frame rate is approximately 50 FPS which is around 5 times the transmission rate in our experiment.

\begin{table}
    \caption{Processing time of SLAM on every frame (example)}
    \begin{center}
    \resizebox{.5\columnwidth}{!}{%
    \begin{tabular}{|c|c|}
    \hline
    \textbf{\textit{frames}} & \textbf{\textit{processing time}} \\
    \hline
    0&1.903s \\
    \hline
    1 &43ms \\
    \hline
    2 &30ms \\
    \hline
    3 &45ms \\
    \hline
    4 &18ms \\
    \hline
    5 &17ms \\
    \hline
    6 &14ms \\
    \hline
    ... & ...... \\
    \hline
    \end{tabular}
    \label{tab4}%
    }
    \end{center}
\end{table}

However, the processing time will increase dramatically when there are many input categories. To accelerate the reconstruction, we allocated a number of GPU resources on Compute Canada to allow the parallel processing of multiple models. We created multiple threads and assigned each category to a thread. In this way, we guarantee the pipeline running in real-time.

\begin{figure}
    \centering
    \subfigure[]{
    \label{Fig. sub.10}
    \includegraphics[width=2.1cm,height = 1.4cm]{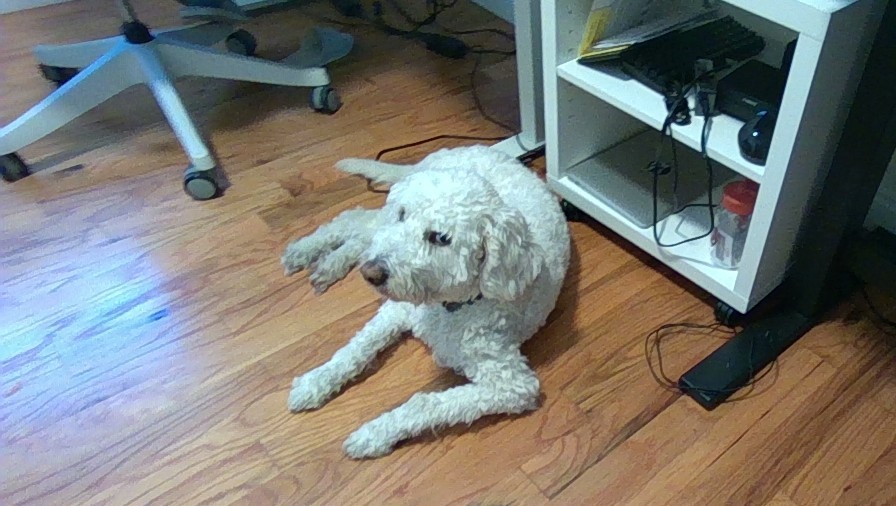}
    }\subfigure[]{
    \label{Fig. sub.11}
    \includegraphics[width=2.1cm,height = 1.4cm]{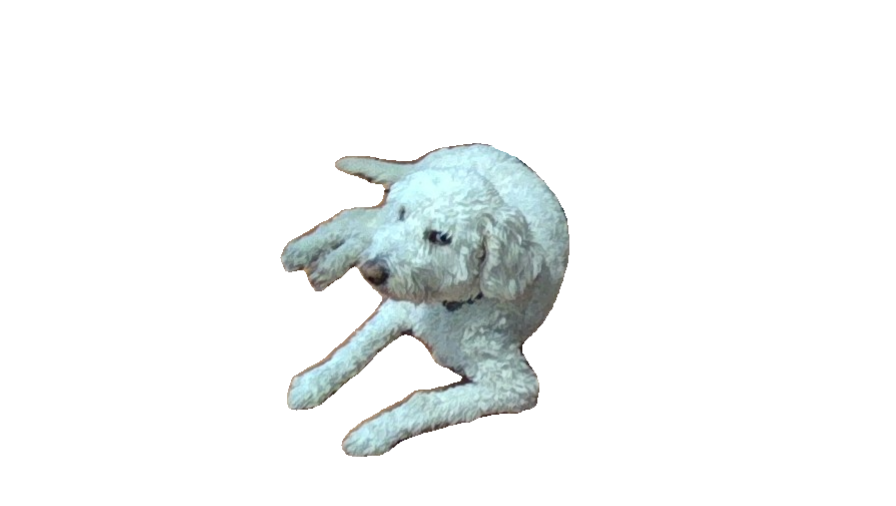}
    }\subfigure[]{
    \label{Fig. sub.12}
    \includegraphics[width=2.1cm,height = 1.4cm]{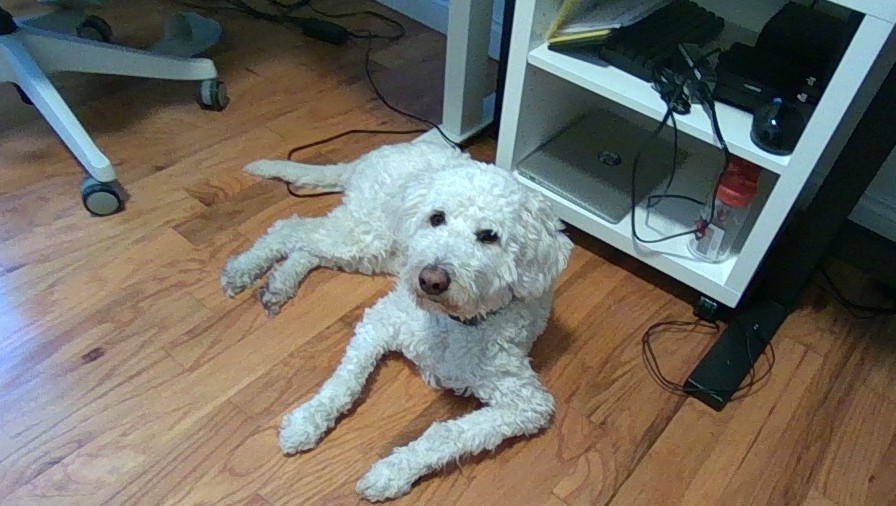}
    }\subfigure[]{
    \label{Fig. sub.13}
    \includegraphics[width=2.1cm,height = 1.4cm]{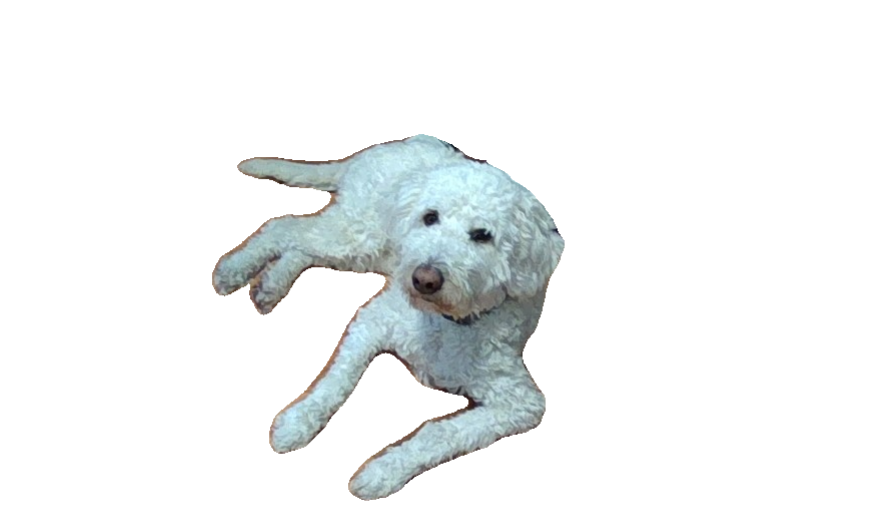}
    }
    \subfigure[]{
    \label{Fig. sub.14}
    \includegraphics[width=2.1cm,height = 1.4cm]{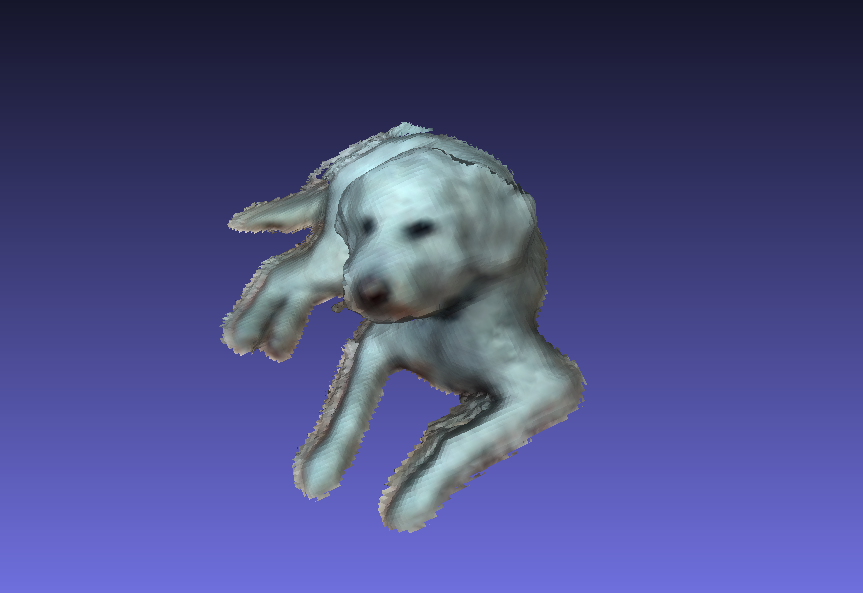}
    }\subfigure[]{
    \label{Fig. sub.15}
    \includegraphics[width=1.8cm,height = 1.4cm]{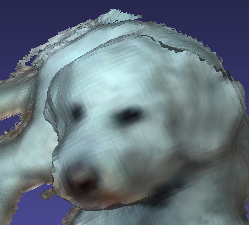}
    }\subfigure[]{
    \label{Fig. sub.16}
    \includegraphics[width=2.1cm,height = 1.4cm]{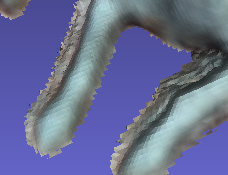}
    }
    \caption{Reconstruction on Customized Datasets}
    \label{fig:reconcustom}
\end{figure}

We validate the reconstruction by the outlooking quality of the 3D models. We tested the reconstruction on the Standford Lounge RGB-D Datasets, which have been the benchmark of SLAM. These datasets were collected in an ideal condition with stable camera movement and high image quality. The reconstructed object appears realistic with details and textures (Figure \ref{fig:3dmodels} and \ref{fig:segmented_pc}). As a comparison, we performed a reconstruction on dynamic objects, such as a dog. When the dog had a slight move (Figure \ref{Fig. sub.10} and \ref{Fig. sub.12}), the reconstructed object was generated with blurs and burrs (Figure \ref{Fig. sub.15} and \ref{Fig. sub.16}) due to the poor registration in frame synthesizing step. Our pipeline currently focuses on the reconstruction of static objects. Dynamic objects from the 2D segmentation are not considered to be candidates for reconstruction in our current implementation.

\section{Conclusion and Future Work}
In this paper, we presented a robust real-time segmentation and reconstruction system leveraging RGB-D images to generate accurate and detailed individual 3D models of objects from a captured scene. This system employs state-of-the-art instance segmentation techniques to perform pixel-level segmentation on the RGB-D data, separating foreground objects from the background and reconstructing them into distinct 3D models. This system runs on a high-performance computation platform. The real-time 3D modelling can be applied in a wide range of domains, including augmented/virtual reality, interior design, urban planning, road assistance, security system and more.

An essential aspect of this project is achieving real-time performance, ensuring that the segmentation and 3D reconstruction can be executed in a continuous and time-sensitive manner. To accomplish this, we proposed an extraction method of consecutive frames to reduce the network load and guarantee the quality of reconstruction at the same time. In the backend, we adopted a multi-process SLAM pipeline to perform the 3D reconstruction in parallel. We executed the clustering on the collection of objects to generate individuals. 

In order to robustly handle the challenging scenarios, we employ the industry-leading object detection framework YOLO for instance segmentation. Some defects of YOLO, such as duplicated or false detection, have been improved to make the object representation more accurate and ensure that the reconstructed models align with the targets. 

In the future, we will employ other real-time instance segmentation models like Mask R-CNN or PointRend to our segmentation tasks to see if we can find a better alternative. Also, we will integrate the edging segmentation model - Segment Anything Model (SAM) to extend the detection of unfamiliar objects without prerequisite training. SAM can also generate high-quality masks \cite{kirillov2023segment}, which potentially eliminates the burr of the masks and enhances the accuracy of our segmentation task.

The current project is focusing on the indoor scene and reconstruction of static objects. In the future, we will extend the application scenario to the outdoors. Moreover, we will develop a scene graph to store all detected objects in the database for urban planning and construction alignment. With more frequent and unstable movement, there will be challenges to the reconstruction of the SLAM pipeline.

\bibliographystyle{IEEEtran}
\bibliography{saa}

\end{document}